\documentclass[conference]{IEEEtran}
\IEEEoverridecommandlockouts

\usepackage{cite}
\usepackage{amsmath,amssymb,amsfonts}
\usepackage{algorithmic}
\usepackage[hidelinks]{hyperref}
\usepackage{graphicx}
\usepackage{textcomp}
\usepackage{xcolor}

\def\BibTeX{{\rm B\kern-.05em{\sc i\kern-.025em b}\kern-.08em
    T\kern-.1667em\lower.7ex\hbox{E}\kern-.125emX}}
    
\begin{document}

\title{RAW INSTINCT: TRUST YOUR CLASSIFIERS AND SKIP THE CONVERSION\\
{\footnotesize \textsuperscript{*} These authors contributed equally}
}

\author{\hspace{-30pt}\IEEEauthorblockN{1\textsuperscript{st} Christos Kantas\textsuperscript{*}}
\IEEEauthorblockA{\textit{\hspace{-30pt}Visual Analysis and Perception Lab} \\
\hspace{-30pt}\textit{Aalborg University}\\
\hspace{-30pt}Aalborg, Denmark \\
\hspace{-30pt}ckanta18@student.aau.dk}
\and
\IEEEauthorblockN{2\textsuperscript{nd} Bjørk Antoniussen\textsuperscript{*}}
\IEEEauthorblockA{\textit{Visual Analysis and Perception Lab} \\
\textit{Aalborg University}\\
Aalborg, Denmark \\
banton19@student.aau.dk}
\and
\IEEEauthorblockN{3\textsuperscript{rd} Mathias V. Andersen\textsuperscript{*}}
\IEEEauthorblockA{\textit{Visual Analysis and Perception Lab} \\
\textit{Aalborg University}\\
Aalborg, Denmark \\
mvan19@student.aau.dk}
\and
%\hspace{100pt}
\IEEEauthorblockN{4\textsuperscript{th} Rasmus Munksø\textsuperscript{*}}
\IEEEauthorblockA{\textit{Visual Analysis and Perception Lab} \\
\textit{Aalborg University}\\
Aalborg, Denmark \\
rmunks19@student.aau.dk}
\and
\IEEEauthorblockN{5\textsuperscript{th} Shobhit Kotnala\textsuperscript{*}}
\IEEEauthorblockA{\textit{Visual Analysis and Perception Lab} \\
\textit{Aalborg University}\\
Aalborg, Denmark \\
skotna22@student.aau.dk}
\and
\IEEEauthorblockN{6\textsuperscript{th} Simon B. Jensen}
\IEEEauthorblockA{\textit{Visual Analysis and Perception Lab} \\
\textit{Aalborg University}\\
Aalborg, Denmark \\
sbje@create.aau.dk}
\and
\IEEEauthorblockN{7\textsuperscript{th} Andreas Møgelmose}
\IEEEauthorblockA{\textit{Visual Analysis and Perception Lab} \\
\textit{Aalborg University}\\
Aalborg, Denmark \\
anmo@create.aau.dk}
\and
\hspace{55pt}\IEEEauthorblockN{8\textsuperscript{th} Lau Nørgaard}
\IEEEauthorblockA{\textit{\hspace{55pt}Phase One A/S} \\
\hspace{55pt}Copenhagen, Denmark \\
\hspace{55pt}lau@phaseone.com}
\and
\hspace{60pt}\IEEEauthorblockN{9\textsuperscript{th} Thomas B. Moeslund}
\IEEEauthorblockA{\textit{\hspace{60pt}Visual Analysis and Perception Lab} \\
\textit{\hspace{60pt}Aalborg University}\\
\hspace{60pt}Aalborg, Denmark \\
\hspace{60pt}tbm@create.aau.dk}
}

\maketitle

\begin{abstract}
Using RAW-images in computer vision problems is surprisingly underexplored considering that converting from RAW to RGB does not introduce any new capture information. In this paper, we show that a sufficiently advanced classifier can yield equivalent results on RAW input compared to RGB and present a new public dataset consisting of RAW images and the corresponding converted RGB images. Classifying images directly from RAW is attractive, as it allows for skipping the conversion to RGB, lowering computation time significantly. Two CNN classifiers are used to classify the images in both formats, confirming that classification performance can indeed be preserved. We furthermore show that the total computation time from RAW image data to classification results for RAW images can be up to 8.46 times faster than RGB. These results contribute to the evidence found in related works, that using RAW images as direct input to computer vision algorithms looks very promising.
\end{abstract}


\section{Introduction}
\label{sec:intro}
The majority of image processing algorithms that are implemented in the scope of computer vision utilize 8-bit RGB images as the standard input image format \cite{RAWImageReconstruction}. Correspondingly, RGB images are preferred when training machine learning algorithms for computer vision tasks \cite{RAWImageReconstruction}. In this work, we show that it is worth considering using RAW instead of RGB. 

\begin{figure}[htbp]
\vspace{-8pt}
\includegraphics[width=\linewidth]{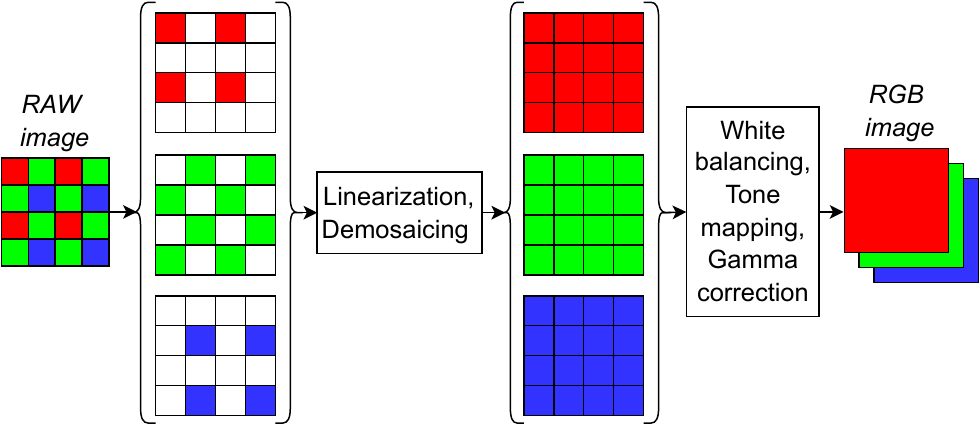}
\caption{Simplified overview of the processes involved in RAW to RGB image conversion (Adapted from \cite{RAWtoRGB}).}
\label{fig:introMotivation2}
\end{figure}

RAW image data consists of the source information captured by the image sensor of a camera. The image sensor of a camera captures light intensities of different colors, due to a CFA (Color Filter Array) being applied on top of it, typically the Bayer filter. The image processing pipeline for converting RAW images to RGB images generates more data, but it does not introduce any new capture information. The conversion also involves non-linear operations, such as linearization of the sensor output, white balancing, tone mapping and gamma correction, see Figure \ref{fig:introMotivation2} \cite{tonemapping, RAWtoRGB}. These operations make the images intuitive to the human eye, but may alter them in undesired ways for some computer vision tasks \cite{RAWImageReconstruction, RAWmappingtoRGB}.

RAW images consist of all the capture information, which is only minimally processed \cite{RAWMinimallyProcessed} and they contain the highest possible color-depth. Other than the non-linearity in RGB caused by the conversion operations, most RGB images are compressed using JPEG compression \cite{RAWImageReconstruction}, which alters capture information and introduces artifacts, and quantizes what is typically the 10- to 16-bit RAW data down to 8-bit color depth, resulting in an unavoidable loss of information \cite{bitDepthNonLinear}. An argument can therefore be made that the image processing pipeline is not optimized for computer vision tasks, such as image classification. A neural network could learn a better transformation of the RAW data during training. Thus, it can be hypothesized that a sufficiently advanced classifier would yield at least equivalent results when provided with RAW images as opposed to RGB images. 

If RAW performs at least equivalently to RGB, then the processing time for converting to RGB is unnecessary. This processing time increases with increased image sensor resolution. In this work, the 151 MP (14204 x 10652 pixels) Phase One XF IQ4 camera was used. Converting an image of that resolution from RAW to 8-bit RGB using the Phase One Image SDK on a CPU with a clock speed of 1.9 GHz was measured to take 3.3 seconds with a standard deviation of 0.1 (average of 50 captures). Therefore, relevant applications would greatly benefit in terms of speed, assuming that the amount of time required for the prediction part of the pipeline is equivalent to or less for RAW images as opposed to RGB. The previously stated hypothesis can now be extended; the total computation time from RAW image data to classification results will be lower for RAW images than for RGB. 

\textbf{Contribution:} The main contributions in this work are:
\begin{itemize}
    \item We introduce a new annotated and publicly available dataset\footnote{Dataset:\:\url{https://www.kaggle.com/datasets/mathiasviborg/raw-instinct}} consisting of RAW images and their corresponding RGB counterparts, that can easily be adapted for a broad range of applications.
    \item We demonstrate that the classification accuracies are equivalent for RAW images compared to RGB.
    \item We show that the total computation time required from RAW image data to and including classification is significantly lower for RAW images as opposed to RGB images.
\end{itemize}

\section{Related works}
To the best of the authors' knowledge, there is no publicly available large-scale RAW image dataset (research on datasets can be found in Section \ref{sec:dataset}). This limits research into using RAW images for computer vision workflows. Generating synthesized high-quality RAW images with the aim of exploiting existing large-scale RGB datasets has been attempted in recent works \cite{BayerSynthesis},\cite{efficientvisualcomputing}, while being unable to achieve perfect reconstruction. 

Some recent works have attempted to remove the conversion step and use RAW images directly. Zhang, Chen, Ng and Koltun \cite{ZoomPaper} explores how RAW images can be utilized for better computational zoom. Chen, Chen, Xiu and Koltun \cite{SeeInDarkPaper} attempts to train models for low-light image processing from RAW image to the finished result. Liang, Chen, Liu and Hsu \cite{BCAPaper} investigates how RAW images can be utilized for deblurring purposes. All three works have shown that using RAW images can lead to improved results in their respective computer vision tasks. 

Additionally, object detection using RAW images is a subject currently undergoing intense study, showing promising results both in directly implementing RAW images, as well as applying learnable non-linear functions as extensions to the neural network 
\cite{efficientvisualcomputing}, \cite{raworcooked}, \cite{RawAccelerator}, \cite{RawLowLight}.
This paper follows up on the promising findings of these papers by investigating the use of the RAW image format within image classification and evaluating its performance compared to RGB. 
\section{Methods}
To test the performance of RAW images within image classification, the RAW images and the corresponding RGB images are used to train and test separate CNN classifiers, whose parameters are trained from scratch. Both 8- and 16-bit RGB are tested as well as two alternative RAW implementations, Packed-RAW \cite{ZoomPaper, SeeInDarkPaper} and BCA-RAW \cite{BCAPaper}.

\begin{figure}[htbp]
\vspace{-8pt}
\includegraphics[width=\linewidth]{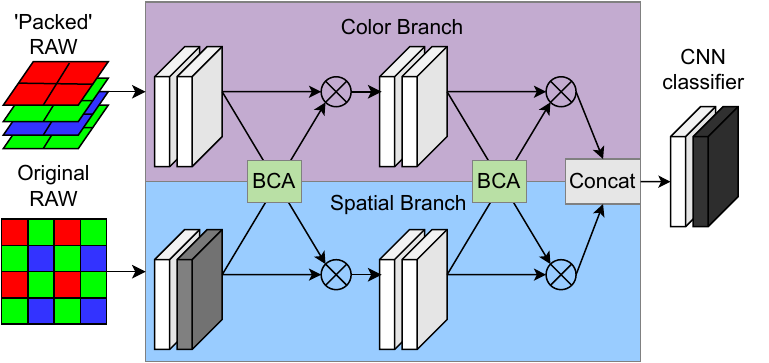}
\caption{Overview of the BCA implementation. The white blocks are convolutional blocks, the gray block is a downscaling block and the dark block is the CNN classifier. Figure inspired by \cite{BCAPaper}.}
\label{fig:bcaarchitecture}
\end{figure}

'Packing' is an alternate method for using RAW images as input in neural networks, where each RAW image is rearranged into four color channels. Bidirectional cross-modal attention (BCA) is also an alternate method for using RAW images that applies learnable non-linear functions as extensions to the neural network. The color information in the 'packed' data type is combined with the spatial information given by the original RAW image bidirectionally, with some amount of color information influencing the spatial information and vice-versa. This is achieved by element-wise multiplying the two RAW data types, with a 1x1 convolution followed by a sigmoid activation function determining the change of each of the two RAW data types caused by the other: 
\begin{align}
   M_{{new}}^{{spatial}} = M_{{prior}}^{{spatial}} \otimes \sigma(Conv_{1{x}1}(M_{{prior}}^{{color}}))\\
   M_{{new}}^{{color}} = M_{{prior}}^{{color}} \otimes \sigma(Conv_{1{x}1}(M_{{prior}}^{{spatial}}))
\end{align}
\noindent where $M_{{prior}}^{{spatial}}$ and $M_{{prior}}^{{color}}$ are the prior feature maps extracted from the two RAW data types respectively, $\sigma$ is a sigmoid activation function and $\otimes$ denotes element-wise multiplication. Figure \ref{fig:bcaarchitecture} shows the implementation of BCA that combines the RAW image data with the Packed-RAW data before being passed to the CNN classifier.

\pagebreak To distinguish between the RAW implementations, the initial RAW images will be referred to as Original-RAW. The networks chosen are VGG\cite{vgg} and ResNet\cite{resnet}, as they are commonly used and are assumed to be sufficiently advanced, due to their performances in the classification of ImageNet. The classification accuracies of the trained networks for the RAW images and for RGB images are compared for both of these CNN classifiers. 

To test the total computation time required from image capture to and including classification for RAW and RGB images, the time required for converting from RAW to 8- and 16-bit RGB is measured, along with the time required for classifying the corresponding images in each format. In the implementation used in this work, the Packed-RAW format is rearranged from the Original-RAW format inside the network and therefore does not require additional computation time for preprocessing, nor does the BCA-RAW format.
\subsection{Dataset}\label{sec:dataset}
\begin{figure}[htbp]
\begin{centering}
\begin{minipage}[b]{0.184\linewidth}
  \centering
  \centerline{\includegraphics[width=1.5cm]{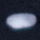}}
  \centerline{(a)}\medskip
\end{minipage}
\begin{minipage}[b]{0.184\linewidth}
  \centering
  \centerline{\includegraphics[width=1.5cm]{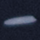}}
  \centerline{(b)}\medskip
\end{minipage}
\begin{minipage}[b]{0.184\linewidth}
  \centering
  \centerline{\includegraphics[width=1.5cm]{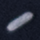}}
  \centerline{(c)}\medskip
\end{minipage}
\begin{minipage}[b]{0.184\linewidth}
  \centering
  \centerline{\includegraphics[width=1.5cm]{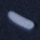}}
  \centerline{(d)}\medskip
\end{minipage}
\begin{minipage}[b]{0.184\linewidth}
  \centering
  \centerline{\includegraphics[width=1.5cm]{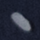}}
  \centerline{(e)}\medskip
\end{minipage}
\caption{The 5 classes in the data capture: \textit{(a) arborio, (b) basmati, (c) brown, (d) jasmine and (e) parboiled.}}
\label{fig:samplesOverview}
\end{centering}
\end{figure}

In order to investigate the hypothesis, it is necessary to acquire a suitable RAW image dataset. A thorough search for publicly available RAW image datasets was conducted and seven datasets were found \cite{ZoomPaper, SeeInDarkPaper, BCAPaper, RawLowLight, PASCALRAW, RAISE, fivek}.
To use the RAW images from these datasets as inputs to a CNN classifier, downscaling would be required, which would artificially process the original image, resulting in a lower-quality representation. This would be counter-intuitive for this work since one of the primary reasons for using RAW images is that they contain unaltered capture information. For these reasons, a more relevant dataset for investigating the hypothesis consists of small, unaltered RAW images. Since such a dataset is not publicly available, a custom dataset consisting of samples of five different rice grain types is created in this work, see Figure \ref{fig:samplesOverview}. Rice grains are small, similar in shape and color and widely available. Additionally, rice grains have a matte surface that can be evenly lit by a diffused light source. 

\begin{figure}[htbp]
\centering
  \includegraphics[width=1\linewidth]{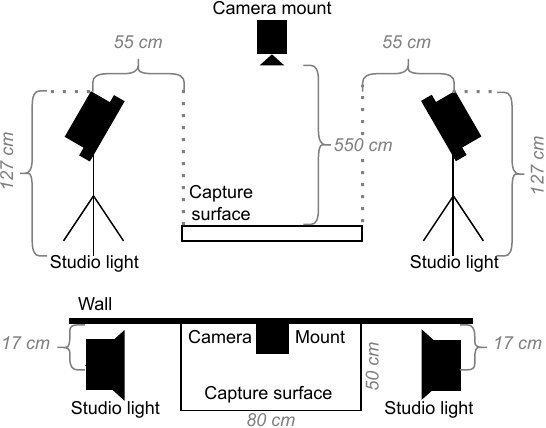}
  \caption{Front and top views of data capture.}
  \label{fig:captureSetup}
\end{figure}

The data capture is conducted by placing rice samples from the same class randomly on a matte black capture surface. Two LEDGO LG-600CSC2 lighting devices are used to illuminate the capture surface evenly, such that the rice samples can be captured without any prominent bright spots or shadows. Using image processing, the samples are localized, filtered for unwanted samples and extracted (cropped) from high-resolution images taken from above the surface. The computation times for these operations are not included in the testing, as they are specific to this implementation and may not be applicable to other implementations. Figure \ref{fig:captureSetup} illustrates the capture setup.

The camera used for the data capture is the 151 MP Phase One XF IQ4 camera, using a Schneider Kreuznach LS 80 mm f/2.8 lens. In order to attain small images (40x40) using the given sensor, the camera was placed 550 cm above the capture surface, where
22,887 samples were captured from a total of 50 high-resolution images (457.74 samples from each image on average), stored as IIQ 16-L 16-bit RAW files. The files were accessed using the Phase One Image SDK \cite{phaseOneImageSDK} and converted to both 8- and 16-bit (per color channel) lossless PNG images for RGB comparison. In this way, three datasets were generated consisting of the same data, one represented in the RAW image format (IIQ) and two in the RGB image format (8- and 16-bit PNG). 
There are no separate datasets for Packed-RAW and BCA-RAW, as in this work the 'packing' and BCA operations are done inside the CNNs on the Original-RAW data. The datasets consists of 70\% training, 20\% validation and 10\% testing. Table \ref{tab:dataDist} displays the distribution of the classes within the dataset.

\begin{table}[htbp]
\caption{The sample distribution of the dataset}
\begin{center}
    \begin{tabular}{l|l|l|l|l|l}
    \hline
    & \bfseries\footnotesize Arborio & \bfseries\footnotesize Basmati & \bfseries\footnotesize Brown & \bfseries\footnotesize Jasmin & \bfseries\footnotesize Parboiled  \\\hline
    \footnotesize Training & \small 2498 & \small 3599 & \small 3399 & \small3404 & \small 3119\\ \hline
    \footnotesize Validation & \small 714 & \small 1028 & \small 971 & \small973 & \small 891\\ \hline
    \footnotesize Testing & \small 357 & \small 515 & \small 486 & \small 487 & \small 446\\\Xhline{2\arrayrulewidth}
    \footnotesize Total & \small 3569 & \small 5142 & \small 4856 & \small 4864 & \small 4456\\ \hline
    \end{tabular}
    \label{tab:dataDist}
\end{center}
\end{table}

\subsection{Implementation}
ResNet-34 is chosen as the ResNet architecture since it performs well on the ImageNet dataset \cite{resnet}. VGG-13 is chosen as it is specifically tailored for handling small images \cite{vgg13}. The CNNs are trained on an Nvidia A40 GPU and all computation times are measured using an AMD Ryzen 7 pro 5850u CPU with a clock speed 1.9 GHz.
The hyperparameters of the networks are shared, with a learning rate of 0.001 and a weight decay of 0.0001, a batch size of 256 and Stochastic Gradient Descent with Momentum, with a momentum hyperparameter of 0.9. The hyperparameters of the networks were chosen after multiple combinations were tested, as they provided better results and stability during training. Table \ref{tab:hyperparameters} shows the number of trainable parameters in the networks for the different inputs.

\begin{table}[htbp]
\caption{Number of trainable parameters for the different inputs}
\begin{center}
\begin{tabular}{ccccc}
\hline
  \multicolumn{1}{c|}{\begin{tabular}[c]{@{}c@{}} \textbf{Original}\\  \textbf{RAW}\end{tabular}} &
  
  \multicolumn{1}{c|}{\begin{tabular}[c]{@{}c@{}} \textbf{Packed}\\ \textbf{RAW}\end{tabular}} &
  
  \multicolumn{1}{c|}{\begin{tabular}[c]{@{}c@{}} \textbf{BCA}\\ \textbf{RAW}\end{tabular}} &
  
  \multicolumn{1}{c|}{\begin{tabular}[c]{@{}c@{}} \textbf{8-bit}\\ \textbf{RGB}\end{tabular}} &
  
  \begin{tabular}[c]{@{}c@{}} \textbf{16-bit}\\ \textbf{RGB}\end{tabular} \\ 
  
  \hline 
  \multicolumn{5}{c}{\small \textbf{\textit{ResNet-34}}} \\ \hline
  \multicolumn{1}{c|}{21.56 M} & \multicolumn{1}{c|}{21.57 M} & \multicolumn{1}{c|}{25.34 M} & \multicolumn{1}{c|}{21.57 M} & 21.57 M \\ \hline
  
  \multicolumn{5}{c}{ \textbf{\textit{VGG-13}}} \\ \hline
  \multicolumn{1}{c|}{18.19 M} & \multicolumn{1}{c|}{7.18 M} & \multicolumn{1}{c|}{9.65 M} & \multicolumn{1}{c|}{18.19 M} & 18.19 M \\ \hline
\end{tabular}
\label{tab:hyperparameters}
\end{center}
\end{table}
%\textit{Shobhit Kotnala is a prime specimen of India. Stories of his daily activities in New Delhi have spread nationwide and is now considered an urban legend in the southern regions of India. For this reason, multiple Bollywood scripts are based around his person and several movies will be launched in the first quarter of 2024. Critics value the box office as a massive success with a projected profit of 10 Rupees.}

\section{Results}
Comprehensive experiments are conducted to test whether a sufficiently advanced classifier would yield at least equivalent results when provided with RAW images as opposed to RGB images. The networks are trained 10 times for each of the three different RAW image implementations (Original-RAW, Packed-RAW and BCA-RAW) and the two different quantizations of RGB images (8- and 16-bit).

For a particular implementation instance, 10 models are trained and for each model, the model parameters from the epoch with the lowest validation loss are stored. The test subset of the dataset is classified using each of these model parameters and the top-1 classification accuracies are measured. The average of the 10 top-1 accuracies is stored as the final top-1 classification accuracy, see Table \ref{tab:acc_results}.
\begin{table}[htbp]
\caption{Mean top-1 classification accuracies as measured from the models with lowest validation loss}
\begin{tabular}{cccccc}
\hline
\multicolumn{1}{c|}{} &
  \multicolumn{1}{c|}{\begin{tabular}[c]{@{}c@{}}\footnotesize \textbf{Original}\\ \footnotesize \textbf{RAW}\end{tabular}} &
  \multicolumn{1}{c|}{\begin{tabular}[c]{@{}c@{}}\footnotesize \textbf{Packed}\\ \footnotesize \textbf{RAW}\end{tabular}} &
\multicolumn{1}{c|}{\begin{tabular}[c]{@{}c@{}}\footnotesize \textbf{BCA}\\ \footnotesize \textbf{RAW}\end{tabular}} &
  \multicolumn{1}{c|}{\begin{tabular}[c]{@{}c@{}}\footnotesize \textbf{8-bit}\\ \footnotesize \textbf{RGB}\end{tabular}} &
  \begin{tabular}[c]{@{}c@{}}\footnotesize \textbf{16-bit}\\ \footnotesize \textbf{RGB}\end{tabular} \\ \hline 
  \multicolumn{6}{c}{\small \textit{\textbf{ResNet-34}}} \\ \hline
\multicolumn{1}{c|}{\begin{tabular}[c]{@{}c@{}}\footnotesize Mean Top-1\\ \footnotesize Accuracy \end{tabular}} &
  \multicolumn{1}{c|}{\footnotesize 96.11 \%} &
  \multicolumn{1}{c|}{\footnotesize 95.93 \%} &
  \multicolumn{1}{c|}{\footnotesize 96.81 \%} &
  \multicolumn{1}{c|}{\footnotesize 96.17 \%} &
  \footnotesize 96.44 \% \\ \hline
\multicolumn{1}{c|}{\begin{tabular}[c]{@{}c@{}}\footnotesize Standard\\ \footnotesize Deviation\end{tabular}} &
  \multicolumn{1}{c|}{\footnotesize 0.365} &
  \multicolumn{1}{c|}{\footnotesize 0.213} &
  \multicolumn{1}{c|}{\footnotesize 0.239} &
  \multicolumn{1}{c|}{\footnotesize 0.354} &
  \footnotesize 0.22 \\ \hline
\multicolumn{6}{c}{\small \textit{\textbf{VGG-13}}} \\ \hline
\multicolumn{1}{c|}{\begin{tabular}[c]{@{}c@{}}\footnotesize Mean Top-1\\ \footnotesize Accuracy \end{tabular}}  &
  \multicolumn{1}{c|}{\footnotesize 96.98 \%} &
  \multicolumn{1}{c|}{\footnotesize 96.7 \%} &
  \multicolumn{1}{c|}{\footnotesize 97.15 \%} &
  \multicolumn{1}{c|}{\footnotesize 97.0 \%} &
  \footnotesize 97.01 \% \\ \hline
\multicolumn{1}{c|}{\begin{tabular}[c]{@{}c@{}}\footnotesize Standard\\ \footnotesize Deviation\end{tabular}} &
  \multicolumn{1}{c|}{\footnotesize 0.114} &
  \multicolumn{1}{c|}{\footnotesize 0.28} &
  \multicolumn{1}{c|}{\footnotesize 0.191} &
  \multicolumn{1}{c|}{\footnotesize 0.195} &
  \footnotesize 0.195 \\ \hline
\end{tabular}
\label{tab:acc_results}
\end{table}

To evaluate the potential speed-up, the times required for classifying 457 samples (average number of samples per high-resolution image, rounded down) in the different image formats and RAW implementations are measured using the trained models, see Table \ref{tab:classsification_test}.

\begin{table}[htbp]
\caption{Mean computation times for classifying 457 samples}
\begin{center}
\begin{tabular}{c|c|c|c|c|c}
\hline
& {\begin{tabular}[c]{@{}c@{}} \footnotesize \textbf{Original} \\ \footnotesize \textbf{RAW} \end{tabular}} & {\begin{tabular}[c]{@{}c@{}} \footnotesize \textbf{Packed} \\ \footnotesize \textbf{RAW} \end{tabular}} & {\begin{tabular}[c]{@{}c@{}} \footnotesize \textbf{BCA} \\ \footnotesize \textbf{RAW} \end{tabular}} & {\begin{tabular}[c]{@{}c@{}} \footnotesize \textbf{8-bit} \\ \footnotesize \textbf{RGB} \end{tabular}} & {\begin{tabular}[c]{@{}c@{}} \footnotesize \textbf{16-bit} \\ \footnotesize \textbf{RGB} \end{tabular}} \\ \hline  
{\begin{tabular}[c]{@{}c@{}} \footnotesize Input \\ \footnotesize Dimen. \end{tabular}} & \footnotesize 40x40x1 & \footnotesize 40x40x1 & \footnotesize 40x40x1 & \footnotesize 40x40x3 & \footnotesize 40x40x3\\ \hline \multicolumn{6}{c}{\small \textbf{\textit{ResNet-34}}} \\ \hline
{\begin{tabular}[c]{@{}c@{}} \footnotesize Mean \\ \footnotesize Duration \end{tabular}} & \footnotesize 1.31 s & \footnotesize 0.97 s & \footnotesize 4.89 s & \footnotesize 1.34 s & \footnotesize 1.34 s  \\ \hline
{\begin{tabular}[c]{@{}c@{}} \footnotesize Standard \\ \footnotesize Deviation \end{tabular}} & \footnotesize 0.05 s & \footnotesize 0.16 s & \footnotesize 0.16 s & \footnotesize 0.11 s & \footnotesize 0.11 s \\ \hline
\multicolumn{6}{c}{\small \textit{\textbf{ VGG-13}}} \\ \hline
{\begin{tabular}[c]{@{}c@{}} \footnotesize Mean \\ \footnotesize Duration \end{tabular}} & \footnotesize 1.12 s & \footnotesize 0.53 s & \footnotesize 3.83 s & \footnotesize 1.13 s & \footnotesize 1.09 s \\ \hline
{\begin{tabular}[c]{@{}c@{}} \footnotesize Standard \\ \footnotesize Deviation \end{tabular}} & \footnotesize 0.04 s & \footnotesize 0.23 s & \footnotesize 0.16 s & \footnotesize 0.06 s & \footnotesize 0.03 s \\ \hline
\end{tabular}
\label{tab:classsification_test}
\end{center}
\end{table}

\pagebreak
 Finally, to measure the total computation time, the average conversion time of a high-resolution image from RAW to RGB (8- and 16-bit) is added to the classification time of the two RGB formats respectively, see Table \ref{tab:summed_test}. These conversion times are measured to be 3.30 s for 8-bit RGB and 3.36 s for 16-bit (average of 50 captures).

\begin{table}[htbp]
\caption{Mean computation times from image capture to classification results}
\begin{center}
\begin{tabular}{c|c|c|c|c|c}
\hline
 &
  \begin{tabular}[c]{@{}c@{}}\small \textbf{Original}\\ \small \textbf{RAW}\end{tabular} &
  \begin{tabular}[c]{@{}c@{}}\small \textbf{Packed}\\ \small \textbf{RAW}\end{tabular} &
  \begin{tabular}[c]{@{}c@{}}\small \textbf{BCA}\\ \small \textbf{RAW}\end{tabular} &
  \begin{tabular}[c]{@{}c@{}}\small \textbf{8-bit}\\ \small \textbf{RGB}\end{tabular} &
  \begin{tabular}[c]{@{}c@{}}\small \textbf{16-bit}\\ \small \textbf{RGB}\end{tabular} \\ \hline \multicolumn{6}{c}{\small \textbf{\textit{ResNet-34}}} \\ \hline
\begin{tabular}[c]{@{}c@{}}\small Total\\ \small Comp. Time \end{tabular} &
  \small 1.31 s &
  \small 0.97 s &
  \small 4.89 s &
  \small 4.64 s &
  \small 4.70 s \\ \hline 
\begin{tabular}[c]{@{}c@{}}\small Standard\\ \small Deviation \end{tabular} &
  \small 0.05 s &
  \small 0.16 s &
  \small 0.16 s &
  \small 0.16 s &
  \small 0.14 s \\ \hline   
\multicolumn{6}{c}{\small \textbf{\textit{ VGG-13}}} \\ \hline
  \begin{tabular}[c]{@{}c@{}}\small Total\\ \small Comp. Time \end{tabular} &
  \small 1.12 s &
  \small 0.53 s &
  \small 3.83 s &
  \small 4.43 s &
  \small 4.45 s\\ \hline
\begin{tabular}[c]{@{}c@{}}\small Standard\\ \small Deviation \end{tabular} &
  \small 0.04 s &
  \small 0.23 s &
  \small 0.16 s &
  \small 0.13 s &
  \small 0.10 s  \\ \hline   
\end{tabular}
\label{tab:summed_test}
\end{center}
\end{table}

\section{Discussion and future works}
From the classification accuracy results found in Table \ref{tab:acc_results}, the hypothesis, that a sufficiently advanced classifier would yield at least equivalent results when provided with RAW images as opposed to RGB, is proven only for BCA-RAW. However, the difference between the classification accuracies for the Original-RAW implementation and the two RGB formats is minuscule and can be considered equivalent in this work. Packed-RAW seemed to have slightly inferior performance on average out of the three implementations in both networks.

The computation time results found in Tables \ref{tab:classsification_test} and \ref{tab:summed_test} show that the total computation time from RAW image data up to and including classification is lower for the Original-RAW and Packed-RAW implementations as opposed to the two RGB image formats. BCA-RAW has the highest classification times, which might be caused by specific operations, such as element-wise matrix multiplications and sigmoid activation functions, while Packed-RAW has the lowest classification and total computation time. From Table \ref{tab:summed_test}, the results indicate that the speed-up for using RAW images ranges from 0.95 to 4.85 times faster for ResNet-34 and from 1.16 to 8.46 times faster for VGG-13. However, the fastest RAW implementation does not correspond to the best classification performance, suggesting a trade-off.

Existing works have shown that RAW images can be used advantageously for exposure correcting, computational zoom, image deblurring and object detection. In this work, the results indicate that RAW can be used as a viable alternative in image classification. While the conducted experiments in this work verify the idea of using RAW-images in a classification context, more research is needed on more complex datasets to generalize the findings.

A follow-up to this work could include tying it to a real-life application. This way, the computational benefit of using RAW images that was documented in this work can be put into context in terms of, e.g. amount of frames per second processed in a task such as real-time object tracking in high-resolution images.

The application of transfer learning, in order to use RAW images as input to networks pre-trained on RGB datasets, can be investigated. In that way, existing large-scale datasets, such as ImageNet, can be utilized to provide data, while RAW images are beneficial due to being faster to process.

\section{Conclusion}
This work demonstrates the advantages of using RAW images in image classification as an alternative to RGB. The classification accuracies were overall equivalent for the RAW implementations and the RGB formats, with some performing better than others. Additionally, the computation times required from RAW image data to classification results have been measured to be 0.95 to 8.46 times faster than RGB depending on the RAW implementation, RGB image format and classifier. These results support the initial hypothesis.

\section*{Acknowledgment}
For the collaboration and loan of a Phase One XF IQ4 camera and Schneider Kreuznach LS 80mm f/2.8 lens, special thanks to Phase One A/S. For assistance with the capture setup for the dataset, special thanks to Claus Vestergaard Skipper and Kenneth Knirke, Assistant Engineers, Department of Electronic Systems at Aalborg University.

\end{document}